\documentclass[11pt]{article}

\usepackage[final]{acl} 

\usepackage[T1]{fontenc}
\usepackage{graphicx}
\usepackage{inconsolata}
\usepackage[utf8]{inputenc}
\usepackage{latexsym}
\usepackage{microtype}
\usepackage{times}

\usepackage{booktabs}
\usepackage{enumitem}
\usepackage{hyperref}
\usepackage{listings}
\usepackage{pifont}
\usepackage{subcaption}
\usepackage[most]{tcolorbox}
\usepackage{url}
\usepackage[table]{xcolor} 
\usepackage{xurl}

\lstdefinestyle{aclpy}{
  language=Python,
  basicstyle=\ttfamily\tiny,
  columns=fullflexible,
  keepspaces=true,
  showstringspaces=false,
  breaklines=true,
  breakatwhitespace=true,
  upquote=true,
  frame=single,
  framerule=0.3pt,
  rulecolor=\color{black!20},
  aboveskip=0.4em,
  belowskip=0.2em,
  xleftmargin=0.4em,
  xrightmargin=0.4em,
  keywordstyle=\color{blue!70!black}\bfseries,      
  commentstyle=\color{green!50!black},              
  stringstyle=\color{red!65!black},                 
  numberstyle=\tiny\color{black!50},
  morekeywords={from,as,with,yield,async,await},
}

\lstdefinestyle{aclout}{
  basicstyle=\ttfamily\tiny,
  columns=fullflexible,
  keepspaces=true,
  showstringspaces=false,
  breaklines=true,
  frame=single,
  framerule=0.3pt,
  rulecolor=\color{black!20},
  aboveskip=0.4em,
  belowskip=0.2em,
  xleftmargin=0.4em,
  xrightmargin=0.4em,
  numbers=none,
  literate={>>>}{{\textcolor{blue!70!black}{>>>}}}3
           {...}{{\textcolor{blue!70!black}{...}}}3
           {\$}{{\textcolor{blue!70!black}{\$}}}1,
}

\newcommand{\cmarkc}{\textcolor{green!60!black}{\ding{52}}}
\newcommand{\xmarkc}{\textcolor{red!70!black}{\ding{56}}}
\newcommand{\umarkc}{\textcolor{blue}{\ding{108}}}

\renewcommand\paragraph[1]{
    \vspace{0.2cm}
    \noindent 
    \textbf{#1}
}

\setlist[itemize]{leftmargin=*, itemsep=2pt, topsep=1pt, parsep=0pt, partopsep=0pt}
\setlist[enumerate]{leftmargin=*, itemsep=2pt, topsep=1pt, parsep=0pt, partopsep=0pt}

\title{
    \raisebox{-0.2\height}{\includegraphics[height=1.2em]{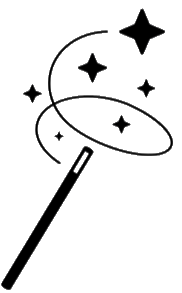}}
    \textsc{Interpreto}: An Explainability Library for Transformers
}

\author{
    Antonin Poché\textsuperscript{1,2,*},
    Thomas Mullor\textsuperscript{1},
    Gabriele Sarti\textsuperscript{3}, 
    Frédéric Boisnard\textsuperscript{4}, \\
    Corentin Friedrich\textsuperscript{1}, 
    Charlotte Claye\textsuperscript{5,6}, 
    François Hoofd\textsuperscript{1,7}, 
    Raphael Bernas\textsuperscript{5}, \\
    Nicholas Asher\textsuperscript{2,8}
    Céline Hudelot\textsuperscript{5}, 
    Fanny Jourdan\textsuperscript{1,*},
\\
\\ 
    \textsuperscript{1}IRT Saint Exupéry Toulouse,
    \textsuperscript{2}IRIT Toulouse,
    \textsuperscript{3}Khoury College of Computer Sciences,\\
    \textsuperscript{4}Ampere,
    \textsuperscript{5}MICS, CentraleSupélec,
    \textsuperscript{6}Scienta Lab,
    \textsuperscript{7}Thales Avionics,
    \textsuperscript{8}ANITI
\\ 
    \textsuperscript{*}\textbf{Equal contribution} \\
    \small{
        \textbf{Correspondence:} antonin.poche | fanny.jourdan @irt-saintexupery.com
    }
}

\begin{document}
\makeatletter
\acl@anonymizefalse
\makeatother
\maketitle

\begin{abstract}
    \textsc{Interpreto} is an open-source Python library for interpreting \textbf{HuggingFace language models}, from early BERT variants to LLMs. It provides two complementary families of methods: \textbf{attribution} methods and \textbf{concept-based} explanations. The library bridges recent research and practical tooling by exposing explanation workflows through a unified API for both \textbf{classification} and text \textbf{generation}. A key differentiator is its end-to-end concept-based pipeline (from activation extraction to concept learning, interpretation, and scoring), which goes beyond feature-level attributions and is uncommon in existing libraries. See \href{https://github.com/FOR-sight-ai/interpreto}{GitHub} or \href{https://for-sight-ai.github.io/interpreto-demo/}{demo website}.
\end{abstract}

\section{Introduction}

Transformer-based language models are widely deployed for classification and text generation, yet understanding their behavior remains important for debugging, bias detection, or safety. Hence, practitioners need tools to simplify access to explanations, including \textbf{attributions} (importance scores assigned to input elements such as tokens or spans) and \textbf{concept-based} explanations (based on higher-level features learned from activations).

Numerous libraries support attributions or concept explanations (Tabs.~\ref{tab:concept-based} and \ref{tab:attribution}). However, these capabilities are often split across packages or tailored to specific modalities or tasks, increasing pipeline complexity. Moreover, documentation or metrics are sometimes missing.

\begin{figure}[t]
        \centering
        \begin{subfigure}{\linewidth}
            \centering
            \includegraphics[width=\linewidth]{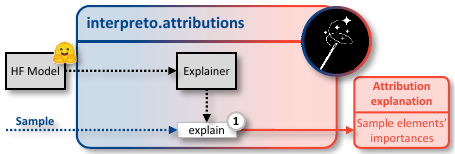}
            \caption{\texttt{\detokenize{attributions}} module (Sec.~\ref{attributions}): a single call explains a sample prediction via element-wise importance scores.}
            \label{fig:attributions}
        \end{subfigure}
        
        \begin{subfigure}{\linewidth}
            \includegraphics[width=\linewidth]{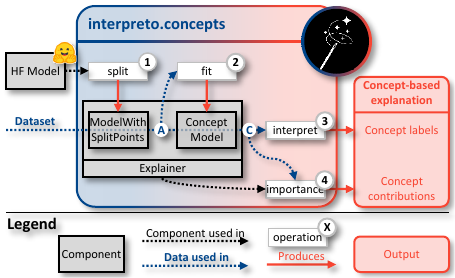}
            \caption{\texttt{\detokenize{concepts}} module (Sec.~\ref{concepts}): \textbf{(1) Split} a HF model and run the dataset to collect latent activations $\textcolor{blue}{A}$; \textbf{(2) Fit} a concept model on $\textcolor{blue}{A}$ and encode activations into concept activations $\textcolor{blue}{C}$; \textbf{(3) Interpret} concepts by linking the dataset with $\textcolor{blue}{C}$ to produce human-readable labels; \textbf{(4) Score} concept contributions by computing importance of $\textcolor{blue}{C}$ w.r.t the model output.}
            \label{fig:concepts}
        \end{subfigure}
        \caption{\textsc{Interpreto} pipelines for HuggingFace (HF) language models. The same API applies to both \textbf{classification} and \textbf{generation} models.}
        \label{fig:pipeline}
        \vspace{-0.3cm}
    \end{figure}

We present \textsc{Interpreto}, an open-source library for interpreting HuggingFace NLP models using both attribution methods and concept-based activation-level analysis (a component of mechanistic interpretability) (Fig.~\ref{fig:pipeline}). \textsc{Interpreto} provides a unified API for classification and text generation, along with visualization tools, metrics, tutorials, and documentation. Our goal is to make research methods easier to adopt and to support benchmarking. We welcome issues and pull requests.

    \paragraph{Availability.}
        \textsc{Interpreto} is open-sourced on \href{https://github.com/FOR-sight-ai/interpreto}{GitHub}\footnote{\scriptsize\url{https://github.com/FOR-sight-ai/interpreto}} under the \textbf{MIT license}. The library comes with its \href{https://for-sight-ai.github.io/interpreto/}{documentation}\footnote{\scriptsize\url{https://for-sight-ai.github.io/interpreto/}} and \href{https://for-sight-ai.github.io/interpreto/notebooks/attribution_walkthrough/}{tutorials}\footnote{\scriptsize\url{https://for-sight-ai.github.io/interpreto/notebooks/attribution_walkthrough/}}. The demonstration \href{https://for-sight-ai.github.io/interpreto-demo/}{website}\footnote{\scriptsize\url{https://for-sight-ai.github.io/interpreto-demo/}} and \href{https://youtu.be/CXep27D3abM}{video}\footnote{\scriptsize\url{https://youtu.be/CXep27D3abM}} can be accessed freely. 

    \paragraph{How to run.}
        \textsc{Interpreto} should be installed via \texttt{uv 
        pip install interpreto}. Examples from Figs.~\ref{fig:att_cls}, \ref{fig:att_gen}, \ref{lst:cpt_gen_code}, and \ref{lst:cpt_cls_code} are runnable end-to-end with only \textsc{Interpreto}; a few optionally use \texttt{datasets} or an external LLM API for labeling.

    \begin{table*}[t]
        \centering
        \small
        \setlength{\tabcolsep}{5pt}
        \begin{tabular}{lcccccc}
            \toprule
                & \textbf{Splitting}    & \textbf{Learning}     & \textbf{Interpretation}       & \textbf{Contributions}    & \textbf{Metrics}  & \textbf{pip \& doc}\\
            \midrule
            \textbf{\href{https://github.com/FOR-sight-ai/interpreto}{\textsc{Interpreto}}}
                & \cmarkc               & \cmarkc               & \cmarkc                       & \cmarkc                   & \cmarkc           & \cmarkc \\
            \midrule
            \textbf{\href{https://nnsight.net/}{NNsight} / \href{https://github.com/Butanium/nnterp/}{NNterp} / \href{https://github.com/TransformerLensOrg/TransformerLens}{TransformerLens}}
                & \cmarkc               & \xmarkc               & \xmarkc                       & \xmarkc                   & \xmarkc           & \cmarkc \\
            \midrule
            \textbf{\href{https://github.com/KempnerInstitute/overcomplete}{Overcomplete}}
                & \umarkc               & \cmarkc               & \umarkc                       & \cmarkc                   & \cmarkc           & \cmarkc \\
            \textbf{\href{https://github.com/jbloomAus/SAELens}{SAELens}}
                & \xmarkc               & \cmarkc*              & \xmarkc                       & \umarkc                   & \cmarkc           & \cmarkc \\
            \textbf{\href{https://github.com/EleutherAI/sparsify}{Sparsify}}
                & \xmarkc               & \cmarkc*              & \xmarkc                       & \umarkc                   & \cmarkc           & \umarkc \\
            \textbf{\href{https://github.com/Prisma-Multimodal/ViT-Prisma}{ViT-Prisma}}
                & \umarkc               & \cmarkc*              & \umarkc                       & \xmarkc                   & \cmarkc           & \xmarkc \\
            \midrule
            \textbf{\href{https://www.neuronpedia.org/}{Neuronpedia}}
                & \cmarkc               & \xmarkc               & \cmarkc                       & \cmarkc                   & \cmarkc           & \umarkc \\
            \textbf{\href{https://github.com/EleutherAI/delphi}{Delphi}}
                & \xmarkc               & \xmarkc               & \cmarkc                       & \xmarkc                   & \cmarkc           & \xmarkc \\
            \textbf{\href{https://github.com/callummcdougall/sae_vis}{SAE-Vis}}
                & \xmarkc               & \xmarkc               & \cmarkc                       & \xmarkc                   & \xmarkc           & \xmarkc \\
            \midrule
            \textbf{\href{https://github.com/adamkarvonen/SAEBench}{SAEBench}}
                & \xmarkc               & \xmarkc               & \cmarkc                       & \xmarkc                   & \cmarkc           & \umarkc \\
            \bottomrule
        \end{tabular}
        \caption{Comparison of libraries for post-hoc unsupervised concept-based. Which steps from Sec.~\ref{concepts} are available? (\cmarkc\ supported; \xmarkc\ not supported; \umarkc\ either not for language models, not directly, or not both; * only SAEs). "pip \& doc" if existing pip package and documentation. We grouped libraries in four categories: 1) Model splitting (model agnostic access to activations); 2) Concepts learning; 3) Interpretations and visualizations; and 4) Benchmarks.}
        \label{tab:concept-based}
        \vspace{-0.3cm}
    \end{table*}
    
    \begin{table}[t]
        \centering
        \small
        \setlength{\tabcolsep}{4pt}
        \begin{tabular}{lccccc}
            \toprule
                & \textbf{Cls} & \textbf{Gen} & \textbf{Metrics} & \textbf{Simple viz} & \textbf{Gran} \\
            \midrule
            \textbf{\href{https://github.com/FOR-sight-ai/interpreto}{\textsc{Interpreto}}}
                & \cmarkc & \cmarkc & \cmarkc & \cmarkc & \cmarkc \\
            \textbf{\href{https://captum.ai/}{Captum}}
                & \cmarkc & \cmarkc & \cmarkc & \xmarkc & \xmarkc \\
            \textbf{\href{https://ferret.readthedocs.io/en/latest/}{Ferret}}
                & \cmarkc & \xmarkc & \cmarkc & \xmarkc & \xmarkc \\
            \textbf{\href{https://inseq.org/en/latest/}{Inseq}}
                & \xmarkc & \cmarkc & \xmarkc & \xmarkc & \xmarkc \\
            \textbf{\href{https://shap.readthedocs.io/en/latest/}{SHAP}}
                & \cmarkc & \cmarkc & \xmarkc & \cmarkc & \xmarkc \\
            \bottomrule
        \end{tabular}
        \caption{Comparison of language-attribution libraries. (\cmarkc\ supported; \xmarkc\ not supported). \textbf{Cls}: sequence classification; \textbf{Gen}: text generation;  \textbf{Metrics} refer to faithfulness metrics; \textbf{Simple viz}: interactive generation visualization; \textbf{Gran}: granularity (token, words, sentences).}
        \label{tab:attribution}
        \vspace{-0.3cm}
    \end{table}

    \paragraph{Contributions.}
        \begin{itemize}
          \item \textbf{\texttt{attribution} module (Sec.~\ref{attributions})} supports HuggingFace models for classification and text generation, with 10 attribution methods, 2 evaluation metrics, and attributions' granularity control.
          \item \textbf{\texttt{concepts} module (Sec.~\ref{concepts}) } provides an end-to-end concept-based pipeline. It wraps \texttt{nnsight} \citep{fiottokaufman2024nnsight} to split HuggingFace language models. It offers 15 concept-learning options (mostly via \texttt{overcomplete} \citep{fel2025overcomplete}), 3 concept-interpretation methods, concept-importance estimation, and 7 metrics.
        \end{itemize}

\section{Related work} \label{related-work}

    For language models' interpretability, open-source libraries fall into two categories: attribution and mechanistic interpretability (MI) libraries. We further detail related libraries in appendix \ref{apx:libraries}.
    
    Tab.~\ref{tab:attribution} compares \textsc{Interpreto} with attribution libraries, including \href{https://shap.readthedocs.io/en/latest/}{SHAP} \citep{lundberg2017unified} and \href{https://captum.ai/}{Captum} \citep{kokhlikyan2020captum}. \textsc{Interpreto} differs by supporting explanations for both classification and text generation, providing evaluation metrics, and exposing granularity controls.

    Tab.~\ref{tab:concept-based} compares \textsc{Interpreto} with MI libraries. These can be grouped into four categories: 1) tooling for model splitting, such as \href{https://nnsight.net/}{NNsight} \citep{fiottokaufman2024nnsight}; 2) concepts learning, such as \href{https://github.com/jbloomAus/SAELens}{SAELens} \citep{bloom2024saelens}; 3) concepts interpretations, such as \href{https://github.com/EleutherAI/delphi}{Delphi} \citep{paulo2025automatically}; and 4) benchmarks, such as \href{https://github.com/adamkarvonen/SAEBench}{SAEBench} \citep{karvonen2025saebench}. \textsc{Interpreto}'s main differentiator is that it integrates these steps into a single package with documentation and examples, reducing the effort needed to apply them in practice.

\section{System overview}

    \paragraph{Environment.}
        \textsc{Interpreto} is available and was tested with python from $3.10$ to $3.13$, torch$>=2.0$, transformers$>=4.22$, and nnsight$>=0.5.1$.
    
    \paragraph{Supported models and tasks.}
        \textsc{Interpreto} supports classic encoders, decoders, and encoder-decoders via the HuggingFace API (tested list in appendix~\ref{apx:tested_architectures}). Concretely, the \texttt{attributions} module supports \texttt{SequenceClassification} and \texttt{CausalLM} models. The \texttt{concepts} module also supports \texttt{MaskedLM} and \texttt{Seq2SeqLM}.

    \paragraph{Workflow and Interface.}
        \textsc{Interpreto} has two modules; their main components are shown in Fig.~\ref{fig:pipeline} and described in Secs~\ref{attributions} and \ref{concepts}. In the \texttt{attributions} module, an explainer takes a model and an input sample and produces token- or word-level importance scores for the model prediction. In the \texttt{concepts} module, the workflow is: i) split model and extract an activation dataset; ii) learn concepts as recurring patterns in these activations; iii) interpret the learned concepts; and iv) estimate their importance for predictions.
    
    \paragraph{Evaluation.}
        To validate correctness,  we provide unit tests for all functions and test more than 15 model architectures (Appendix~\ref{apx:tested_architectures}). We also include sanity checks on manually constructed models in both tasks to verify the correctness of explanations. Finally, \citet{mussot2026bringing} used \textsc{Interpreto} in an industrial setting, supporting its accessibility and practical usefulness.

\section{Demonstration}

    \paragraph{Content.}
        The demonstration material is a \href{https://for-sight-ai.github.io/interpreto-demo/}{website} that serves as a gallery of \textsc{Interpreto} explanations. Appendix~\ref{apx:demo} is a screenshot of the website. Each explanation includes a button that opens a minimal, runnable snippet.

    \paragraph{Requirements.}
        The demonstration website only requires an internet connection, since explanations are precomputed. The snippets are end-to-end runnable with \textsc{Interpreto} alone, with optional dependencies for data loading (\texttt{datasets}) and concept labeling (external LLM API).

    \paragraph{Types of explanations.}
        The available explanations, interactions, and visuals depend on the task, the explanation family, and user-specified settings. The gallery covers: i) \textbf{Attributions for classification}: examples that explain all classes in parallel or only the predicted class; ii)\textbf{Attributions for generation}: users select an output element and view the corresponding attributions; iii) \textbf{Concepts for classification}: global concept-based explanations, either general across classes or learned class-wise; and iv) \textbf{Concepts for generation}: local concept-based analyses for individual generations.


    \paragraph{Website interface.}
        All explanations are interactive: depending on the view, users can select classes, output tokens, or concepts. Users can choose the task (classification/generation), model, dataset, explanation family (attribution/concepts), method subset, and the instance to inspect.

    \paragraph{Models and datasets.}
        All models are hosted on HuggingFace; generating explanations requires a GPU (see computation times). The gallery covers 6 models: 3 classifiers (DistilBERT/IMDB \citep{sanh2019distilbert,maas2011learning}, BERT/emotion \citep{devlin2019bert,saravia2018carer}, RoBERTa/AG-News \citep{liu2019roberta,zhang2015character}) and 3 generators (GPT-2 \citep{radford2019language}, Qwen3-0.6B \citep{yang2025qwen3}, Llama~3.1~8B \citep{grattafiori2024llama}).

    \paragraph{User scenario.}
        A practitioner debugs an emotion recognition model. They first explore the explanations on the demo website. From which, they copy a runnable code snippet for KernelSHAP attributions and adapt it for their use case. Although highlighted tokens appear plausible, predictions remain incorrect in several cases. They then switch to class-wise concept-based explanations in the gallery and reproduce the concept discovery step locally (e.g., Semi-NMF). Inspecting the global concepts' importance suggests that ``fear'' and ``sadness'' are not well separated; further checks reveal an inconsistent class-index mapping across data sources, explaining the observed errors.


    \paragraph{Computation times.}
        Website explanations are precomputed. Attributions typically require 10--100 forward passes or 5--20 gradient computations (seconds). Concept pipelines are dominated by activation extraction and concept importance scoring; small runs take minutes on an RTX~3080 (Fig.~\ref{lst:cpt_gen_code}), while large SAEs can take hours.

\section{\texttt{interpreto.attributions}} \label{attributions}

\begin{figure}[t]
        \centering
        
        \begin{subfigure}[t]{\linewidth}
            \centering
            \lstset{style=aclpy}
            \begin{lstlisting}
import torch
from transformers import AutoTokenizer, AutoModelForSequenceClassification
from interpreto import Lime, plot_attributions

# Load the model and tokenizer
repo_id = "nateraw/bert-base-uncased-emotion"
model = AutoModelForSequenceClassification\
    .from_pretrained(repo_id)
tokenizer = AutoTokenizer.from_pretrained(repo_id)
classes_names =\
    ['sadness', 'joy', 'love', 'anger', 'fear', 'surprise']

# Instantiate the explainer
explainer = Lime(model, tokenizer)

# Compute the explanation
attributions = explainer(
    "We are thrilled to present you Interpreto!",
    targets=torch.arange(len(classes_names)).view(1, -1)
)

# Visualize the explanation
plot_attributions(attributions[0], classes_names=classes_names)\end{lstlisting}
        \end{subfigure}
        \hfill
        \begin{subfigure}[t]{\linewidth}
            \centering
            \includegraphics[width=\linewidth]{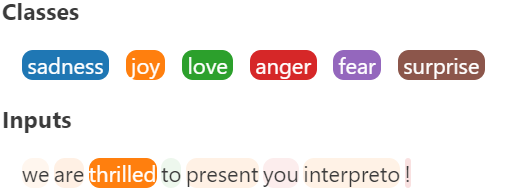}
        \end{subfigure}
        \caption{Classification attribution explanation with \textsc{Interpreto}. (Top) Minimal code to compute attributions with \texttt{Lime} \citep{ribeiro2016should} for a BERT model \citep{devlin2019bert} fine-tuned on emotion recognition \citep{saravia2018carer}. (Bottom) Notebook visualization of the resulting attributions: words are highlighted by importance, and colors correspond to classes. The model predicts the ``joy'' emotion, mainly driven by ``thrilled''.}
        \label{fig:att_cls}
        \vspace{-0.3cm}
    \end{figure}

\begin{figure}[t]
        \centering
        
        \begin{subfigure}[t]{\linewidth}
            \centering
            \lstset{style=aclpy}
            \begin{lstlisting}
from transformers import AutoTokenizer, AutoModelForCausalLM
from interpreto import Occlusion, plot_attributions

# Load the model and tokenizer
repo_id = "Qwen/Qwen3-0.6B"
model = AutoModelForCausalLM.from_pretrained(repo_id)
tokenizer = AutoTokenizer.from_pretrained(repo_id)

# Instantiate the explainer
explainer = Occlusion(model, tokenizer)

# Compute the explanation
attributions = explainer.explain(
    "An interpretability open-source library is great,",
    " but, for HF language models, it is even better.",
)

# Visualize the explanation
plot_attributions(attributions[0])\end{lstlisting}
        \end{subfigure}
        \hfill
        \begin{subfigure}[t]{\linewidth}
            \centering
            \centering
            \includegraphics[width=\linewidth]{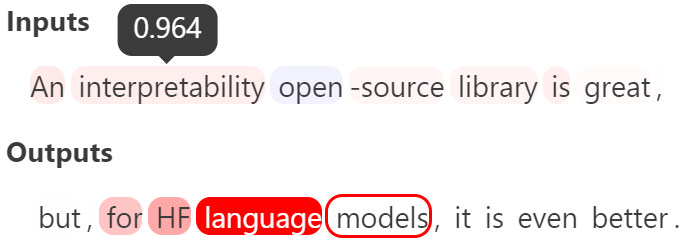}
        \end{subfigure}
        \caption{Generation attribution explanation with \textsc{Interpreto}. (Top) Minimal code to compute attributions with \texttt{Occlusion} \citep{zeiler2014visualizing} for a Qwen3-0.6B causal language model \citep{yang2025qwen3} on a hand-crafted input-output pair. (Bottom) Notebook visualization: in generation, each output token is a separate prediction, so attributions usually form an input-output matrix; instead, \textsc{Interpreto} lets users select a token of interest. Here, the token ``models'' is selected (red border), and the most influential input token is ``language''.}
        \label{fig:att_gen}
        \vspace{-0.3cm}
    \end{figure}

    \paragraph{Definition and taxonomy.}
        Attribution explains a prediction by estimating the contribution of input features. In vision, this yields heatmaps; in NLP, it highlights salient tokens or words (Figs.~\ref{fig:att_cls} and \ref{fig:att_gen}). Methods fall into two broad families. \textbf{Perturbation-based} approaches modify the input and measure the effect on the output. \textbf{Gradient-based} approaches use derivatives of the output with respect to the input. These families are closely related: gradients correspond to the limit of infinitesimal perturbations.

    \paragraph{API.}
        With \textsc{Interpreto} (as in other attribution libraries listed in appendix \ref{attr_lib}) the workflow has three steps (summarized in Fig.~\ref{fig:attributions} and illustrated with code examples in Figs.~\ref{fig:att_cls} and \ref{fig:att_gen}):
        \begin{enumerate}
            \item Instantiate an \texttt{AttributionExplainer} with a HuggingFace model and tokenizer.
            \item Compute explanations for \texttt{inputs} (the text to explain), optionally providing \texttt{targets} that specify what to explain (class indices for classification and selected output tokens for generation).
            \item Visualize the resulting attributions.
        \end{enumerate}

        We recommend that users generate the output text they want to explain and pass it to \textsc{Interpreto} as \texttt{targets}.

    \paragraph{Additional features.}
    In addition to the ten methods above, several attribution explainers' parameters provide useful variants:
    \begin{itemize}
        \item \texttt{granularity}: Explanations can be produced at the token, word, or sentence level. Because subword tokens and special tokens can be difficult to interpret, the default is \textbf{word}-level.
        \item \texttt{input\_x\_gradient}: For gradient-based methods, returning the elementwise product of input and gradient is formally justified \citep{shrikumar2017learning}; enabled by default.
        \item \texttt{inference\_mode}: Choose the output space (logits, softmax, or log softmax). Since softmax couples classes, the default is logits.
    \end{itemize}

    \paragraph{Available methods and metrics.}
        \begin{itemize}
            \item \textbf{Perturbation-based (4)}: KernelSHAP \citep{lundberg2017unified}, LIME \citep{ribeiro2016should}, Occlusion \citep{zeiler2014visualizing}, and Sobol \citep{fel2021look}.
            \item \textbf{Gradient-based (6)}: GradientSHAP \citep{lundberg2017unified}, Integrated Gradients \citep{sundararajan2017axiomatic}, Saliency \citep{simonyan2014deep}, SmoothGrad \citep{smilkov2017smoothgrad}, SquareGrad \citep{hooker2019benchmark}, and VarGrad \citep{hooker2018benchmark}.
            \item \textbf{Faithfulness metrics (2)}: Insertion, Deletion \citep{petsiuk2018rise}. 
        \end{itemize}
    
    \paragraph{Custom method.}
    \textsc{Interpreto} is designed to make new methods lightweight to implement: users write the method-specific computation, while the library handles shared engineering concerns. When existing building blocks are close to the desired behavior, we recommend reusing them and adapting an existing implementation. Internally, the pipeline follows three stages (similar in spirit to the abstraction of \citet{ferre2025muppet}):
    \begin{itemize}
        \item \textbf{Perturbations:} starting from one sample, construct perturbed variants (used by most methods, including some gradient-based ones). For a custom perturbation-based method, users can inherit from the input-IDs perturbator and implement the \texttt{mask\_function} method; for gradient-based methods, the corresponding component is the embedding perturbator.
        \item \textbf{Inference or gradients:} run forward and/or backward passes on perturbed samples. This stage typically only needs subclassing to modify the forward or backward computation.
        \item \textbf{Aggregation:} aggregate intermediate scores into element-wise importance scores. For a custom aggregator, users inherit from the base aggregator and implement \texttt{aggregate}.
    \end{itemize}

\section{\texttt{interpreto.concepts}} \label{concepts}
    
    \paragraph{Definition and taxonomy.}
        Concepts are interpretable computational features used by the model (examples in Figs.~\ref{fig:cpt_gen_out} and \ref{fig:cpt_cls_out}). If users can i) predict a sample’s concept activations from the input and ii) predict the output from these activations, they can simulate (and understand) the model’s behavior. There are two axes \citet{bhalla2024towards}:
        \begin{itemize}
            \item \textbf{Post hoc vs.\ by design:} concepts can be introduced during model construction or extracted from a trained model (\textsc{Interpreto} positioning).
            \item \textbf{Supervised vs.\ unsupervised:} Concepts can be specified in advance (\textit{e.g.}, probes \citep{alain2016understanding,belinkov2022probing} or CAVs \citep{kim2018interpretability}) (planned for later) or discovered from activations. \textsc{Interpreto} currently emphasizes unsupervised discovery (dictionary learning).
        \end{itemize}
        Concept-based explanations can be \textbf{global} (concepts that are important for a class) or \textbf{local} (concepts that are active for a specific input or important for the prediction). \textsc{Interpreto} supports both; global analyses often provide the basis for local ones. This line of work is connected to mechanistic interpretability (MI); sparse autoencoders (SAEs) are a common approach for concept discovery.
    
\begin{figure}[t]
        \centering
        \includegraphics[width=\linewidth]{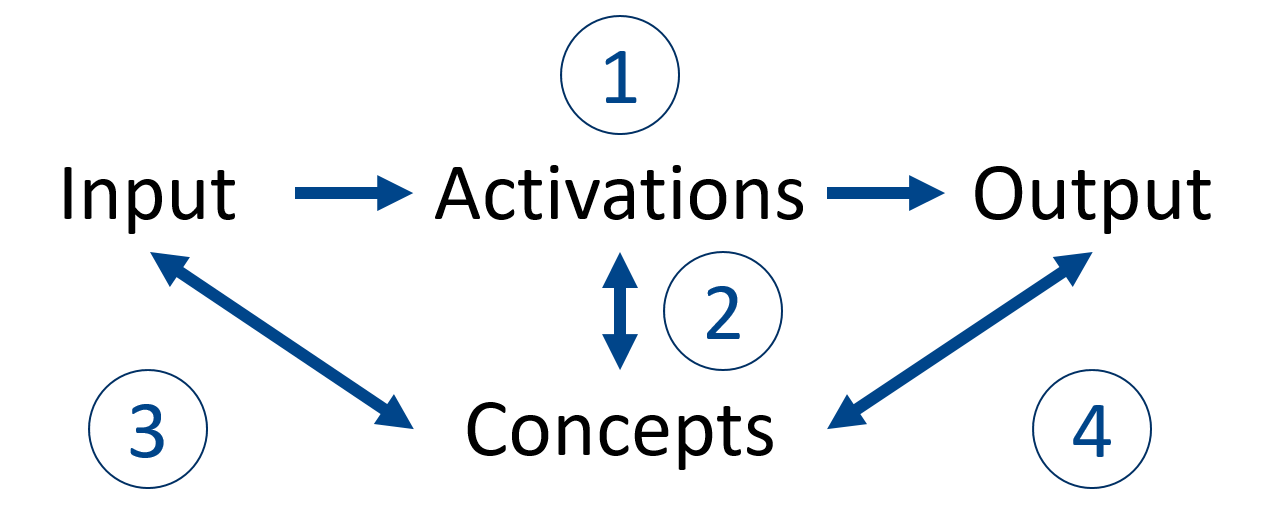}
        \caption{Post-hoc, unsupervised concept-based pipeline. (1) Split the model and extract a dataset of activations. (2) Learn concepts as patterns in these activations. (3) Interpret concepts by assigning human-meaningful labels. (4) Estimate each concept’s importance for the model predictions (Sec.~\ref{cpt_pipeline}).}
        \label{fig:cpt_pipeline}
        \vspace{-0.5cm}
    \end{figure}

    \paragraph{Pipeline.} \label{cpt_pipeline}
        Post-hoc, unsupervised concept methods typically follow four steps \citep{fel2023holistic,poche2025consim}. These steps are reflected in Figs.~\ref{lst:cpt_gen_code} and \ref{lst:cpt_cls_code} and summarized in Fig.~\ref{fig:cpt_pipeline}:
        \begin{enumerate}
            \item \textbf{Split the model} into a \textbf{feature extractor} and a \textbf{predictor}, then run a dataset through the extractor to collect activations.
            \item \textbf{Construct the concept space:} train a concept model on these activations. Each learned concept corresponds to one dimension of the concept space. The concept model maps between the activation space and the concept space.
            \item \textbf{Interpret concepts:} assign human-meaningful labels (\textit{e.g.}, by linking inputs to concept activations) to each concept dimension.
            \item \textbf{Estimate concept importance:} quantify each concept’s contribution to the predictions.
        \end{enumerate}

    \paragraph{Classification vs. generation.}
        The API is shared across tasks, but recommended settings differ:
        \begin{itemize}
            \item \textbf{Activation granularity:} use the [CLS] token for classification and tokens for generation. Granularity is set during activation extraction.
            \item \textbf{Interpretation:} match the interpretation granularity to the activation granularity; the tutorials provide task-specific guidance.
        \end{itemize}
    
    \paragraph{Methods to define the concept space.}
        \begin{itemize}
            \item \textbf{Neurons as concepts:} treat individual neurons as concepts \citep{geva2022transformer}.
            \item \textbf{Dictionary learning:} KMeans \citep{ghorbani2019towards}, PCA \citep{zhang2021invertible}, SVD  \citep{graziani2023concept,jourdan2023taco}, ICA \citep{poche2025consim}, NMF \citep{zhang2021invertible,fel2023craft,jourdan2023cockatiel}, Semi-NMF, Convex NMF \citep{fel2025archetypal}.
            \item \textbf{Sparse autoencoders (SAEs):} Vanilla SAE \citep{bricken2023monosemanticity,huben2024sparse}, Jump-ReLU SAE \citep{rajamanoharan2024jumping}, Top-$k$ SAE, Batch Top-$k$ \citep{gao2025scaling}, Matching Pursuit SAE \citep{costa2025flat}.
        \end{itemize}

    \paragraph{Methods for concept interpretation.}
        Top-$k$ vocabulary tokens \citep{geva2022transformer}, top-$k$ activating examples/words/n-grams (MaxAct) \citep{bricken2023monosemanticity}, and LLM-based labeling \citep{bills2023language}.

    \paragraph{Methods to estimate concept importance.}
        Concept-to-output gradients \citep{fel2023holistic} and concept\,$\times$\,gradients \citep{poche2025consim}. Compatibility with the attributions module is planned.

    \paragraph{Metrics.}
        Evaluate \textbf{concept-space} faithfulness via MSE \citep{bricken2023monosemanticity} and FID \citep{fel2023holistic}; measure sparsity \citep{bricken2023monosemanticity} and Stability \citep{fel2023holistic,paulo2025sparse}; finally, evaluate explanation \textbf{general usefulness} via ConSim \citep{poche2025consim}.

    \paragraph{Custom method.}
        Each step is modular, allowing users to mix and match: split points, concept models, and interpretation methods. To add a new method, users implement the corresponding abstraction and focus on the computational core; the library handles integration and execution. For new components, we recommend using existing implementations as templates.

    \begin{figure}[!t]
        \centering
        \lstset{style=aclpy}
        \begin{lstlisting}
import datasets
from transformers import AutoModelForCausalLM

from interpreto import ModelWithSplitPoints
from interpreto.concepts import SemiNMFConcepts
from interpreto.concepts.interpretations import LLMLabels
from interpreto.model_wrapping import llm_interface

TOKEN = ModelWithSplitPoints.activation_granularities.TOKEN

# 1.1 Split your model in two parts
mwsp = ModelWithSplitPoints(
    "Qwen/Qwen3-0.6B",
    automodel=AutoModelForCausalLM,
    split_points=[5],
)

# 1.2 Compute a dataset of activations
dataset = datasets.load_dataset(
    "fancyzhx/ag_news")["train"]["text"][:100]
activations = mwsp.get_activations(dataset, TOKEN)

# 2. Train the concept model
explainer = SemiNMFConcepts(mwsp, nb_concepts=20)
explainer.fit(activations)

# 3. Interpret the concepts
llm = llm_interface.OpenAILLM(os.getenv("OPENAI_API_KEY"))
interpreter = LLMLabels(
    concept_explainer=explainer,
    activation_granularity=TOKEN,
    llm_interface=llm,
    k_context=10,
)
interpretations = interpreter.interpret(
    "all", dataset, activations)\end{lstlisting}
        \caption{Generation concept-based code example (steps 1-3). We analyze Qwen3-0.6B \citep{yang2025qwen3} by learning concepts with Semi-NMF \citep{fel2025archetypal} from activations computed on 100 AG-News samples \citep{zhang2015character}. We then label concepts with GPT-4.1-nano \citep{openai2025introducing} using our default prompting scheme. We omit concept importance, which is less informative for global analyses of generation models. It runs in under 3 minutes on an RTX 3080 (10GB).}
        \label{lst:cpt_gen_code}
        \vspace{-0.3cm}
    \end{figure}

    \begin{figure}[!t]
        \centering
        \lstset{style=aclout}
        \begin{lstlisting}
Proper Noun Indicators
Entity Names
Market Focus
Abbreviation Markers
Named Entity Indicators
Keyword Indicators
Acronym Prefixes
Entity References
Proper Nouns or Named Entities
Quantification or comparison markers\end{lstlisting}
        \caption{First ten concept labels produced by the LLM-based interpreter for the code in Fig.~\ref{lst:cpt_gen_code}. Labels depend on the system prompt; by default, we request short but discriminative labels. Note that the example uses an API to externalize computations, but the labeling can also be done locally. Additionally, concept-labels are hard to tune; they can be vague and redundant or too precise and lengthy; this is discussed in the limitations section \ref{limitations}.}
        \label{fig:cpt_gen_out}
        \vspace{-0.2cm}
    \end{figure}
    
    \paragraph{API.}
        The API follows the pipeline steps. Fig.~\ref{fig:concepts} summarizes the module, and Figs.~\ref{lst:cpt_gen_code} and \ref{lst:cpt_cls_code} provide minimal examples; additional examples are available in the \href{https://for-sight-ai.github.io/interpreto/notebooks/classification_concept_tutorial/}{classification} and \href{https://for-sight-ai.github.io/interpreto/notebooks/generation_concept_tutorial/}{generation} tutorials.
        
        \begin{enumerate}
            \item \textbf{Split the model} wrap the HuggingFace model with \texttt{ModelWithSplitPoints} (built on \texttt{nnsight} \citep{fiottokaufman2024nnsight}). Collect activations with the wrapper.
            \item \textbf{Construct the concept space:} Collect activations through the split model wrapper. Instantiate an explainer (linked to concept models) with the wrapped model and fit on the activation dataset. Many explainers leverage \texttt{overcomplete} \citep{fel2025overcomplete}.
            \item \textbf{Interpret concepts:} run an interpretation method on the fitted explainer.
            \item \textbf{Estimate concept importance:} compute concept importance scores (\textit{e.g.}, via the explainer’s concept-to-output gradient method).
        \end{enumerate}

\section{Limitations and outlook} \label{limitations}

    \textsc{Interpreto} focuses on HuggingFace text language models and does not cover all interpretability settings. Some limitations are inherent to the explainability methods, and we leave improving the state of the art to other work, including the plausibility-faithfulness trade-off and human biases in interpretation. SAE-based methods can be computationally expensive, and LLM-based interpretations are sensitive to prompt design. Additionally, the library is under active development, so we plan to expand its coverage. Nonetheless, some elements will remain outside the library's scope. Details are provided below.

    \paragraph{Limitations of the literature}
        \begin{itemize}[leftmargin=10pt]
            \item There is no single method to govern them all. Users have to apply several methods and compare them using metrics.
            \item Explanations are subject to human biases (such as confirmation bias and over-interpretations), therefore, they should be treated carefully.
            \item The meaning of attribution scores depends on the methods. Similar scores from different methods can mean different things.
            \item LLM-based concepts interpretations are highly sensitive to the given prompt. Depending on the specifications, labels may be too general, preventing us from differentiating among several concepts. Or too specific to be actionable.
            \item A not interpretable concept can come from a bad model, a bad concept-space, or a bad interpretation, and we lack ways to determine which one.
        \end{itemize}

    \paragraph{Outlook (ongoing development).}
        \begin{itemize}[leftmargin=10pt]
            \item Add supervised concept options (CAVs \citep{kim2018interpretability} and probes \citep{alain2016understanding,belinkov2022probing}).
            \item Extend interpretation-specific metrics (\textit{e.g.}, Detection/Fuzzing/Clarity/Purity \citep{paulo2025automatically,puri2025fade}).
            \item Add additional attribution methods (\textit{e.g.}, RISE \citep{petsiuk2018rise}, RFEM \citep{ayyar2025there}) and metrics (\textit{e.g.}, AOPC comprehensiveness and sufficiency).
            \item Link the two modules to allow input-to-concepts attributions.
        \end{itemize}

    \paragraph{Outlook (later).}
        We plan to extend library coverage to ViT (Vision Transformer) to later aim for multi-modal transformer support.

    \paragraph{Out of scope.}
        We do not plan to include circuit-level MI methods, data attribution, or feature visualization; the library will remain centered on HuggingFace model workflows.

\section*{Ethical statement}

    \paragraph{Impact.}
        \textsc{Interpreto} lowers the barrier to applying attribution and concept-based interpretability methods to HuggingFace language models. Easier access can support model auditing, debugging, and documentation, including the identification of biases and recurring failure modes. At the same time, interpretability outputs can be misread as faithful causal explanations; results depend on method choice, hyperparameters, and presentation. Users should therefore treat explanations as diagnostic evidence rather than ground truth, and corroborate findings with additional tests (e.g., counterfactual checks, ablations, and evaluation on targeted slices).
    
    \paragraph{LLM use.}
        LLM-assisted code suggestions were used during development. Any generated code was manually reviewed by the developer and again during pull-request review before merging. During paper writing, LLMs were only used for editing existing text (e.g., sentence rewriting for clarity) and for flagging missing elements expected in a system demonstration paper; no manuscript text was generated from scratch.

\section*{Acknowledgments}  
    
    Our work has benefited from the AI Cluster ANITI and the research programs DEEL\footnote{\url{https://www.deel.ai/}} and FOR\footnote{\url{https://www.irt-saintexupery.com/for-program/}}. ANITI is funded by the France 2030 program under the Grant agreement n°ANR-23-IACL-0002. DEEL and FOR are integrative programs of the AI Cluster ANITI, designed and operated jointly with IRT Saint Exupéry, with the financial support from its industrial and academic partners and the France 2030 program under the Grant agreement n°ANR-10-AIRT-01.

    We also acknowledge the support of Pr. Philippe Muller and Dr. Grégory Flandin.

\bibliography{biblio}

\clearpage
\appendix
\section{Related libraries details} \label{apx:libraries}

    \paragraph{Attribution libraries.} \label{attr_lib}
        Explainability libraries can be grouped into two broad families: attribution and mechanistic/concept ones. In the first family, \href{https://github.com/marcotcr/lime}{LIME} \citep{ribeiro2016should} and \href{https://shap.readthedocs.io/en/latest/}{SHAP} \citep{lundberg2017unified} are model- and modality-agnostic toolkits. For NLP, \href{https://inseq.org/en/latest/}{Inseq} \citep{sarti2023inseq} and \href{https://ferret.readthedocs.io/en/latest/}{Ferret} \citep{attanasio2023ferret} provide language-focused APIs and visualizations. \href{https://captum.ai/}{Captum} \citep{kokhlikyan2020captum} offers a broad set of attribution methods in PyTorch but does not include unsupervised concept discovery. \href{https://deel-ai.github.io/xplique/latest/}{Xplique} \citep{fel2022xplique} spans both attribution and concept tools, with an emphasis on vision rather than NLP.
    
    \paragraph{Mechanistic interpretability libraries.} \label{mi_lib}
        Concept-based explanations are part of mechanistic interpretability (MI). MI libraries target MI researchers and specific stages of the concept pipeline (see Sec.~\ref{cpt_pipeline}). For activation access and intervention, see \href{https://github.com/TransformerLensOrg/TransformerLens}{TransformerLens} \citep{nanda2022transformerlens}, \href{https://nnsight.net/}{NNsight} \citep{fiottokaufman2024nnsight}, \href{https://www.neuronpedia.org/}{Neuronpedia} \citep{lin2023neuronpedia}, and \href{https://github.com/Butanium/nnterp/}{NNterp} \citep{dumas2025nnterp}. For training concept models (e.g., SAEs), \href{https://github.com/EleutherAI/sparsify}{Sparsify} \citep{belrose2025sparsify}, \href{https://github.com/jbloomAus/SAELens}{SAELens} \citep{bloom2024saelens}, \href{https://github.com/KempnerInstitute/overcomplete}{Overcomplete} \citep{fel2025overcomplete}, and \href{https://github.com/Prisma-Multimodal/ViT-Prisma}{ViT-Prisma} \citep{joseph2025prisma} provide tooling. For interpretation and visualization, \href{https://github.com/EleutherAI/delphi}{Delphi} \citep{paulo2025automatically} and \href{https://github.com/callummcdougall/sae_vis}{SAE-Vis} \citep{mcdougall2024sae} are complementary. Finally, \href{https://github.com/adamkarvonen/SAEBench}{SAEBench} \citep{karvonen2025saebench}, \href{https://github.com/stanfordnlp/axbench}{AxBench} \citep{wu2025axbench}, and \href{http://pyvene.ai/}{Pyvene} \citep{wu2024pyvene} aim to evaluate concept-based methods, including via steering.

\section{Tested models architectures} \label{apx:tested_architectures}

    We test more than 15 model architectures: Albert \citep{lan2020albert}; BART \citep{lewis2020bart}; BERT \cite{devlin2019bert}; DistilBERT \citep{sanh2019distilbert}; Electra \citep{clark2020electra}; Roberta \citep{liu2019roberta}; T5 \citep{raffel2020exploring}; GPT2 \citep{radford2019language}; GPT-Neo \citep{black2021gptneo}; GPT-J \citep{black2021gptneo}; CodeGen \citep{nijkamp2023codegen}; Falcon \citep{almazrouei2023falcon}; Llama3 \citep{grattafiori2024llama}; Mistral \citep{jiang2023clip}; Starcoder \citep{lozhkov2024starcoder}; Qwen3 \citep{yang2025qwen3}.

\section{Demonstration website example} \label{apx:demo}

\begin{figure}[t]

    \vspace{-0.3cm}
    \centering
    \includegraphics[width=\linewidth]{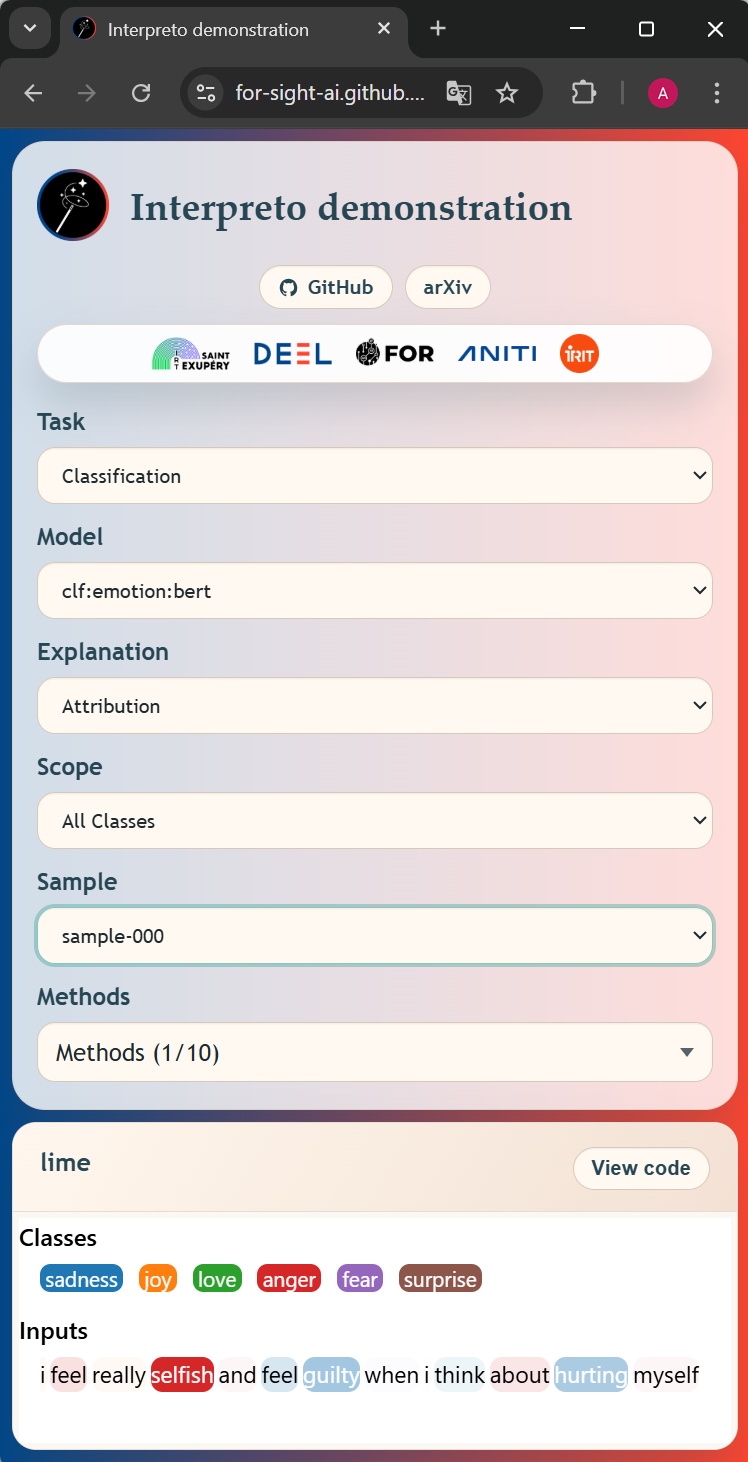}
    \caption{Screenshot from the demonstration website. The user can select a task, model, and explanation format to explore the explanation gallery. \href{https://for-sight-ai.github.io/interpreto-demo/}{Website link}.}
    \label{fig:demo}
    \vspace{-0.4cm}
\end{figure}

\section{Concept-based classification example} \label{apx:cpt_cls}

    \begin{figure*}[t]
        \centering
        
        \begin{subfigure}[t]{\linewidth}
            \centering
            \lstset{style=aclpy}
            \begin{lstlisting}
import os
from datasets import load_dataset
from transformers import AutoModelForSequenceClassification
from interpreto import ModelWithSplitPoints
from interpreto.concepts import LLMLabels, SemiNMFConcepts
from interpreto.model_wrapping.llm_interface import OpenAILLM

# set the granularity to [CLS] to keep only this token's activation
granularity = ModelWithSplitPoints.activation_granularities.CLS_TOKEN
# set the LLM interface used to generate labels based on the constructed prompts
llm_interface = OpenAILLM(api_key=os.getenv("OPENAI_API_KEY"), model="gpt-4.1-nano")

# ---------------------------------------------------------------------------------------------------------------------------
# 1. Split the model into two and get a class-wise dataset of activations
# 1.1 Split the model
model_with_split_points = ModelWithSplitPoints(
    "textattack/distilbert-base-uncased-ag-news", split_points=[5],  # split at the sixth layer
    automodel=AutoModelForSequenceClassification, device_map="cuda", batch_size=1024
)
# 1.2 Construct the dataset of activations (extract the ones related to the class)
# load the AG-News dataset (here we use only 1000 examples to go faster, but the more, the better)
inputs = load_dataset("fancyzhx/ag_news")["train"]["text"][:1000]
# Compute the [CLS] token activations
activations = model_with_split_points.get_activations(inputs, granularity, include_predicted_classes=True)

for target, class_name in enumerate(classes_names):  # iterate over classes
    # 1.3 Take the subset for the target class (predicted class)
    indices = (activations["predictions"] == target).nonzero(as_tuple=True)[0]
    class_wise_inputs = [inputs[i] for i in indices]
    class_wise_activations = {k: v[indices] for k, v in activations.items()}

    # -----------------------------------------------------------------------------------------------------------------------
    # 2. train concept model
    concept_explainer = SemiNMFConcepts(model_with_split_points, nb_concepts=20, device="cuda")
    concept_explainer.fit(class_wise_activations)

    # -----------------------------------------------------------------------------------------------------------------------
    # 4. compute concepts importance (before interpretations to limit the number of concepts interpreted)
    gradients = concept_explainer.concept_output_gradient(class_wise_inputs, [target],
                                                          activation_granularity=granularity, batch_size=64)
    # stack gradients on samples and average them over samples
    concept_importances = torch.stack(gradients, axis=0).squeeze().abs().mean(dim=0)  # (num_concepts,)
    # for each class, sort the importance scores
    important_concept_indices = torch.argsort(concept_importances, descending=True).tolist()

    # -----------------------------------------------------------------------------------------------------------------------
    # 3. interpret the important concepts concepts
    llm_labels_method = LLMLabels(concept_explainer=concept_explainer, activation_granularity=granularity,
                                  llm_interface=llm_interface, k_examples=20,)
    concept_interpretations = llm_labels_method.interpret(inputs=class_wise_inputs,
                                                          concepts_indices=important_concept_indices)\end{lstlisting}
            \caption{Classification concept-based explanation code example. We explain a DistilBERT \citep{sanh2019distilbert} classifier fine-tuned on AG News \citep{zhang2015character}. We decompose the concepts with the Semi-NMF \citep{fel2025archetypal} and interpret them with the LLM labels \citep{bills2023language} from GPT-4.1-nano \citep{openai2025introducing}. The concepts are computed for each predicted class.}
            \label{lst:cpt_cls_code}
        \end{subfigure}
        \hfill
        \begin{subfigure}[t]{\linewidth}
            \lstset{style=aclout}
            \begin{lstlisting}
Class: World
	importance: 0.088,	Diverse topics unified by factual reporting and event-focused language
	importance: 0.086,	Consistent focus on named entities, events, and quotations.
	importance: 0.084,	Consistently presents factual info with emphasis on notable events or figures.
	importance: 0.082,	Recurring topics and formal reporting language.
	importance: 0.078,	Concise reporting style with specific names, events, and dates.

Class: Sports
	importance: 0.112,	Patterns involve sports, achievements, and event summaries with emphasis on names, scores...
	importance: 0.105,	Structured sports and news summaries emphasize game events, scores, and highlights, often...
	importance: 0.082,	Focus on key entities, events, and record-breaking achievements.
	importance: 0.076,	Consistent references to sports events, players, and results.
	importance: 0.07,	Concise patterns: factual summaries with focus on specific event outcomes.

Class: Business
	importance: 0.092,	Concise patterns involve financial, infrastructural, and cultural references.
	importance: 0.083,	Structured information emphasis, financial and infrastructural keywords.
	importance: 0.081,	Topical focus, political or economic narratives, and authoritative tone.
	importance: 0.074,	Concise patterns include focus on current events, economic issues, and industry developments.
	importance: 0.072,	Focus on comparative categories and rankings across regions.

Class: Sci/Tech
	importance: 0.121,	Language features that explicitly reference entities, events, or dates.
	importance: 0.101,	Consistent focus on environmental impacts, species, and conservation initiatives.
	importance: 0.095,	Scientific observations of natural phenomena and technological descriptions.
	importance: 0.093,	Factual reporting with scientific, environmental, and technological focus, structured as...
	importance: 0.081,	Factual, technical, and headline-like language with specific details\end{lstlisting}
            \centering
            \caption{List of class-wise concept labels with their global importance for the class.}
            \label{fig:cpt_cls_out}
        \end{subfigure}
        
        \caption{Code and output for a global classification concept-based explanations. Here, concepts are computed class-wise; it is not mandatory, but it gives better, more specific concepts. The model is a DistilBERT \citep{sanh2019distilbert} classifier fine-tuned on the AG News \citep{zhang2015character} dataset. The code and outputs are extracted from the \href{https://for-sight-ai.github.io/interpreto/notebooks/classification_concept_tutorial/}{classification concepts tutorial}.}
        \label{fig:cpt_cls}
        \vspace{-0.5cm}
    \end{figure*}

\end{document}